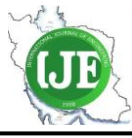

# International Journal of Engineering

Journal Homepage: www.ije.ir

# Kinematic Synthesis of Parallel Manipulator via Neural Network Approach

J. Ghasemi, R. Moradinezhad, M. A. Hosseini*

*Faculty of Engineering & Technology, University of Mazandaran, Babolsar, Iran*

*A B S T R A C T*

In this research, Artificial Neural Networks (ANNs) have been used as a powerful tool to solve the inverse kinematic equations of a parallel robot. For this purpose, we have developed the kinematic equations of a Tricept parallel kinematic mechanism with two rotational and one translational degrees of freedom (DoF). Using the analytical method, the inverse kinematic equations are solved for specific trajectory, and used as inputs for the applied ANNs. The results of both applied networks (Multi-Layer Perceptron and Redial Basis Function) satisfied the required performance in solving complex inverse kinematics with proper accuracy and speed.

*doi*: 10.5829/ije.2017.30.09c.04

## 1. INTRODUCTION

Parallel manipulators have some advantages over other serial compeers such as low inertia, high stiffness, high load carrying capacity and high precision [1]. Tricept Parallel Kinematic Machine tool (PKM), has both rotational and orientational degrees of freedom which makes it one of the famous parallel manipulators used in machining industries[2]. Among the recent researches, Pond and Corretero [2] performed a comparison study among some similar parallel mechanism with the same degrees of freedom. They demonstrated the superiorities of Tricept with respect to the dexterity and workspace volume. Hosseini and Daniali [3-5] introduce weighted factor method to normalize the Jacobian matrix. They use this for optimizing the dexterous workspace shape and size. Also, they illustrated that the optimized structure of Tricept is completely different from other machines that are commonly used by other manufacturers. Further, Hosseini and Daniali [6] suggested Cartesian homogenized Jacobian matrix to evaluate derived performance indices for Tricept PKM.

On the other hand, computing the kinematic formulas on robot's computer is a time and memory consuming process and remarkably decreases the functionality of the robot. In many cases, kinematic calculations cannot perform on real time. Hence, researchers tend to use system identification algorithms such as Artificial Neural Networks (ANNs) to simulate the calculations of the robotic systems [7-9]. Specifically, in parallel mechanisms, researchers try to design a model that is able to provide proper outputs when receives the proper robot input data. These types of models have several applications in control design [10-14]. For example, from recent works, Xu and Li [10] design a neural network to estimate the forward kinematics of a 3-PRS (prismatic-revolute-spherical) parallel manipulator. Also, Parikh and Lam [11] implemented an iterative neural network to a flight simulation system parallel manipulator to solve its forward kinematics problem. Li and Wang [12] suggested that in a 2-DoF (degree of freedom) redundantly actuated parallel robot, applying an appropriate neural network PID controller could improve the trajectory tracking performance and reduce the errors of the system. Guan et al. [13] use neural networks to design a hybrid computational intelligent method for the kinematic analysis of the parallel machine tool. Morella et al. [15] proposed a support vector machine to solve the forward kinematics problems of a parallel manipulator called Stewart platform. In this paper, two models are proposed to estimate the PKM Tricept kinematic equations. The first one is based on the Multi-Layer Perceptron (MLP) NNs and the second one benefit from Radial Basis Function (RBF) NNs.

*Corresponding Author's Email: ma.hosseini@umz.ac.ir (M. A. Hosseini)
[2] www.pkmtricept.com

The rest of the paper is organized as follows. Section 2 introduces the PKM Tricept and its kinematic equations. Section 3 introduces the neural networks. It also provides some basic information about MLP and RBF networks. Section 4 discusses the structure of two NN-based simulations and their results: first, information about configuration of the models is provided, and then results on normalized and real data[3] are provided, respectively. Finally, section 5 includes the concluding remarks.

## 2. TRICEPT PKM AND KINEMATIC EQUATIONS

As shown in Figure 1, the manipulator consists of base platform, moving platform, three active legs and one passive leg. Active legs are linear (prismatic) actuators which connect the base to the moving platform by universal (or spherical) and spherical joints. The passive leg consists of two parts; the upper part, which is a link with constant connected to the moving platform by a spherical joint; and the lower part, which is a prismatic joint, linked to the base and upper parts by a passive universal joint. Moving and global frame, {P(uvw)} and {O(xyz)} are attached to the moving and base platform, respectively.

The geometric model of the $i^{th}$ leg of the Tricept is depicted in Figure 1. The closure equation for this leg can be written as:

$$c + R(a_i + d) = b_i n_{b_i} + l_i + n_{l_i} \qquad (1)$$

where $c$ and $d$ are the vectors from $O$ to $C$ and $C$ to $P$, respectively. $R$ is rotation matrix carrying frame {P} into an orientation coincident with that of frame {O}; $a_i$ is the position vector from $P$ to $A_i$ in frame {P}; $b_i$ is the position vector of point $B_i$ in the global frame. Moreover, $n_{b_i}$ and $n_{l_i}$ are the unit vectors showing the directions of vectors $b_i$ and $l_i$, respectively.

Dot-multiplying both sides of Equation (1) by $n_{l_i}$, upon simplifications leads to:

$$c^T n_{l_i} + (R(a_i + d))^T n_{l_i} - b_i n_{b_i}^T n_{l_i} = l_i \qquad (2)$$

Rewriting Equation (2) for $i = 1, 2, 3$ leads to three quadratic equations which can be solved either numerically or theoretically.

---
[3] Term "real data" means that all numbers are used directly from database without normalizing them.

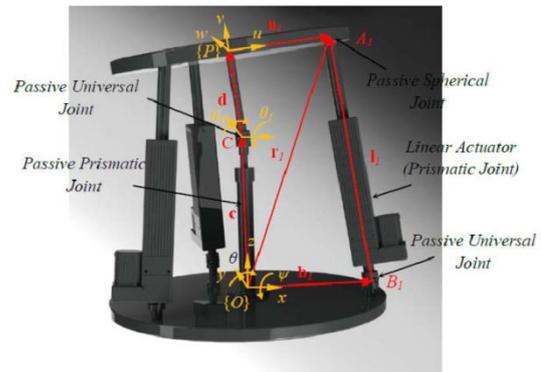

**Figure 1.** Optimum Tricept PKM with maximum dexterous workspace

Siciliano [16] developed the kinematics and studied the manipulability of the Tricept. Pond and Corretero [2] formulated its square dimensionally homogeneous Jacobian matrices by improving the method proposed by Kim and Ryu [17]. Architectural optimization of the Tricept and similar mechanisms was undertaken by Wang and Gosselin [18]. Hosseini and Daniali [19] recalled kinematic equations and solved the equations analytically. Furthermore, Hosseini and Daniali [6] investigated the dexterous workspace and the shape of the mechanism and optimized it with considered constraints.

## 3. MLP AND RBF NEURAL NETWORK

Artificial Neural Network (ANN) is a mathematical model of biological neural networks of the human brain. Using the cells called "Neurons", neural network processes information and makes decisions. Every single neuron is connected to many other neurons and they transmit electrical signals via synapses. The same idea is used in computer science, which tiny interconnected units have been designed to transmit signals to each other [20].

A Multi-Layer Perceptron (MLP) is a feed-forward artificial neural network model consisting of an input layer, one or some hidden layers and an output layer, with each layer fully connected to the next one. Figure 2 shows a typical MLP network.

The linear functions are only able to map an input to one (or some) outputs, without making any reasonable relationship between them. While, using some non-linear functions, called *"Activation Function"*, the MLP networks are able to draw a relationship between input and output data. The Tansig and Logsig are two famous sigmoid functions that are used frequently as MLP networks activation functions. Their equations are represented below, respectively.

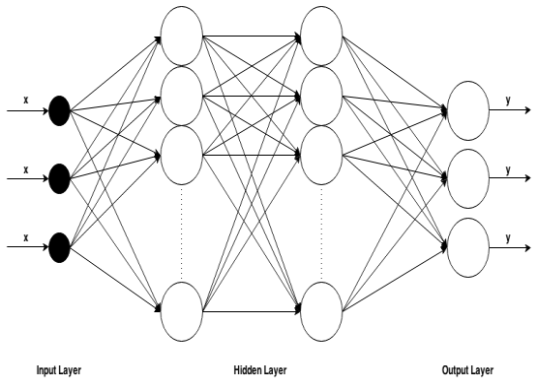

**Figure 2.** Scheme of the MLP network

$$\varnothing(v_i) = \tanh(v_i) \tag{3}$$

$$\varnothing(v_i) = (1 + e^{-v_i})^{-1} \tag{4}$$

As explained, the MLP networks are trained by using nonlinear algorithms to back propagate errors between expected outputs (targets) and network outputs [15]. Selecting the features of the networks is the most challenging part of the network design since choosing the right architecture is more dependent to user's experience than a specific theoretical theorem [10].

Radial Basis Function (RBF) networks are feed-forward artificial neural networks which use the radial basis functions as activation functions. They are typically trained using a supervised training algorithm. Their architecture contains a single hidden layer. Figure 3 shows a standard RBF network.

A scalar function of the input vector, $\varphi: R^n \to R$, is the output of RBF network. By considering $N$ as number of the neurons, $C_i$ as the center vector of neuron $i$, and $a_i$ as the weight of neuron $i$, the equation is:

$$\varphi(x) = \sum_{i=1}^{N} a_i \rho(\|x - c_i\|) \tag{5}$$

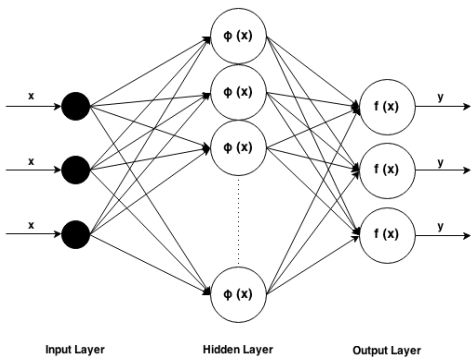

**Figure 3.** Scheme of the standard RBF network

Commonly, the radial basis function is a Gaussian function:

$$\rho(\|x - c_i\|) = \exp\left[-\beta \|x - c_i\|^2\right] \tag{6}$$

A Gaussian basis function is local to the center vector. It is shown by following equation:

$$\lim_{\|x\| \to \infty} \rho(\|x - c_i\|) = 0 \tag{7}$$

This means that making a change in the parameters of one neuron has a negligible effect on input values that are not close to the center of that neuron.

The MLP and RBF have many applications; for instance, function approximation, time series prediction, classification, and system control [12-14, 21-25]. The RBF networks usually train much faster than a typical MLP network. They are also less vulnerable to difficulties with unreliable inputs [20].

## 4. SIMULATION AND RESULTS

The purpose of the inverse kinematic is to find the actuator lengths, given the desired position of the moving platform. Considering Figure 1, the inverse kinematic can be depicted as following, in which the columns of the matrix are the position vectors of the spherical joints [26]:

$$A = \begin{bmatrix} C\theta & S\psi S\theta & C\psi S\theta \\ 0 & C\psi & -S\psi \\ -S\theta & C\theta S\psi & C\theta C\psi \end{bmatrix} \begin{bmatrix} \frac{a}{\sqrt{3}} & -\frac{a}{2\sqrt{3}} & -\frac{a}{2\sqrt{3}} \\ 0 & \frac{a}{2} & -\frac{a}{2} \\ d & d & d \end{bmatrix} - \begin{bmatrix} 0 & 0 & 0 \\ 0 & 0 & 0 \\ c & c & c \end{bmatrix} =$$

$$\begin{bmatrix} \frac{a}{\sqrt{3}} C\theta + dC\psi S\theta & -\frac{a}{2\sqrt{3}} C\theta + \frac{a}{2} S\psi S\theta + dC\psi S\theta & -\frac{a}{2\sqrt{3}} C\theta - \frac{a}{2} S\psi S\theta + dC\psi S\theta \\ -dS\psi & \frac{a}{2} C\psi - dS\psi & -\frac{a}{2} C\psi - dS\psi \\ -\frac{a}{\sqrt{3}} S\theta + dC\theta C\psi + c & \frac{a}{\sqrt{3}} S\theta + \frac{a}{2} C\theta C\psi + dC\theta C\psi + c & \frac{a}{2\sqrt{3}} S\theta - \frac{a}{2} C\theta S\psi + dC\theta C\psi + c \end{bmatrix} \tag{8}$$

$C$ and $S$ stand for the *Cos* and *Sin* functions, $a$ is the moving platforms radius and $c$ is the length of the passive prismatic actuator of the middle limb.

Matrix $B$ groups the position vectors of the universal joints:

$$B = \begin{bmatrix} \frac{b}{\sqrt{3}} & -\frac{b}{2\sqrt{3}} & -\frac{b}{2\sqrt{3}} \\ 0 & \frac{b}{2} & -\frac{b}{2} \\ 0 & 0 & 0 \end{bmatrix} \tag{9}$$

where $b$ indicates the radius of the base platform.

Using Equations (8) and (9), the actuator lengths can be calculated:

$$q_1^2 = \frac{a^2}{3} + \frac{b^2}{3} + c^2 + d^2 - \frac{2}{3} abC\theta + 2cdC\theta C\psi - \frac{2bd}{\sqrt{3}} C\psi S\theta \tag{10}$$

$$q_2^2 = \frac{a^2}{3} + \frac{b^2}{3} + c^2 + d^2 - \frac{1}{2}ab\left(\frac{1}{3}C\theta - \frac{1}{\sqrt{3}}S\psi S\theta + C\psi\right) + \ldots$$
$$+ bd\left(\frac{C\psi S\theta}{\sqrt{3}} + S\psi\right) + 2cdC\theta C\psi - ac\left(\frac{S\theta}{\sqrt{3}} + C\theta S\psi\right) \quad (11)$$

$$q_3^2 = \frac{a^2}{3} + \frac{b^2}{3} + c^2 + d^2 - \frac{1}{2}ab\left(\frac{1}{3}C\theta + \frac{1}{\sqrt{3}}S\psi S\theta + C\psi\right) + \ldots$$
$$+ bd\left(\frac{C\psi S\theta}{\sqrt{3}} - S\psi\right) + 2cdC\theta C\psi - ac\left(\frac{S\theta}{\sqrt{3}} - C\theta S\psi\right) \quad (12)$$

Therefore, our purpose is to design a model based on ANNs, which takes $\theta, \psi$ and $C$ as its inputs and estimates $q_1$, $q_2$ and $q_3$ as its outputs. Figure 4 represents the distribution of input data such as $\theta, \psi$ and $C$.

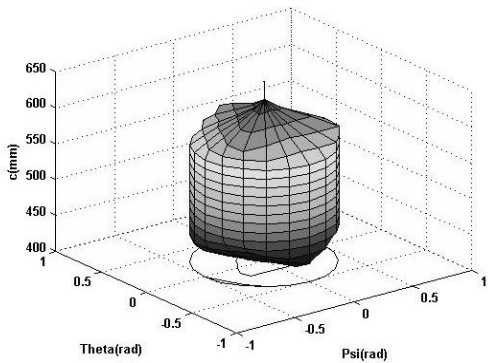

**Figure 4.** The distribution of input data

**TABLE 1.** Statistical information of input and target real data

|  | Min | Max | Mean | Median | Variance |
|---|---|---|---|---|---|
| $\theta$ | -0.5027 | 0.5027 | -6e-4 | 0 | 0 |
| $\psi$ | -0.5027 | 0.5027 | 0.0117 | 0 | 0 |
| $C$ | 426 | 634 | 525.1945 | 530 | 3082.9 |
| $q_1$ | 470.2868 | 664.9327 | 562.3404 | 566.0176 | 2.6868 |
| $q_2$ | 470.2886 | 6649422 | 562.3567 | 566.0284 | 2.6869 |
| $q_3$ | 470.2886 | 664.9422 | 562.321 | 565.9819 | 26.86.7 |

**TABLE 2.** Statistical information of input and target normalized data

|  | Min | Max | Mean | Median | Variance |
|---|---|---|---|---|---|
| $\theta$ | 0 | 1 | 0.4994 | 0.5 | 0.0385 |
| $\psi$ | 0 | 1 | 0.5117 | 0.5 | 0.042 |
| $C$ | 0 | 1 | 0.5077 | 0.5294 | 0.0631 |
| $q_1$ | 0 | 1 | 0.5021 | 0.52 | 0.0633 |
| $q_2$ | 0 | 1 | 0.5022 | 0.52 | 0.0633 |
| $q_3$ | 0 | 1 | 0.502 | 0.5198 | 0.0633 |

Network input and target data are represented in Table 1 and Table 2, respectivly. Scales for $\theta$ and $\psi$ are in radian. $C$, $q_1$, $q_2$ and $q_3$ are represented in millimeters.

Input and target data are stored in two 3-column matrixes, each containing 4818 samples. As even in the most accurate conditions, manipulators are not expected to perform in less than micrometers scale, obtaining a performance with errors less than 1e-3 can be interpreted as an ideal performance. We call this, "**goal error**".

**4. 1. Configuration of Network Parameters**
By analyzing the relationship between number of layers and neurons, and also the speed and the error rate of the simulation, proposed MLP network has been designed with one layer including 5 neurons. The numbers of neurons are selected to effectively cover the size of input data. Performance measure is mean squared error (MSE) and Levenberg-Maquardt is chosen as the training function. The network is set to stop at 222[nd] iteration.

Like MLP network, by analyzing the problem, proposed RBF network is set to use maximum number of 20 neurons (iterations). Since real data have more anomaly than normalized data, the network works with spread value of 200 for real data and 2 for normalized data. Similarly, the performance measure is mean squared error (MSE).

Performance of both networks is evaluated for both normalized and real data.

**4. 2. Performance on Normalized Data**    Data normalization represents all the input and target data in a specific range and it is expected to ease the work of network to approximate the existing relation. In this work, all input and target data are mapped in the interval of [0, 1].

Figure 5 represents the performance of the MLP and RBF network for normalized data. Errors in MLP network rapidly decreases in first 10 epochs. Then, it continues to decrease with a mild slope. It has a breakpoint at 52[nd] epoch, and then continues its progress with a slope bowed to 1. Best performance occurs at the 222[nd] epoch and that is the point that the network stops training. Final MSE is 1.6e-9 that surely satisfies the goal error.

In the RBF network, performance has a sharp slope in first and second epochs but it continues to progress with a mild slope until 15[th] epoch. In that stage, the breakpoint happens and improves the performance significantly. Training the network for more times, it can be shown that using more than 20 neurons does not make any considerable change in the performance of the network. In addition, as the MSE value gets smaller than the goal error, there is no need for more than 20 neurons.

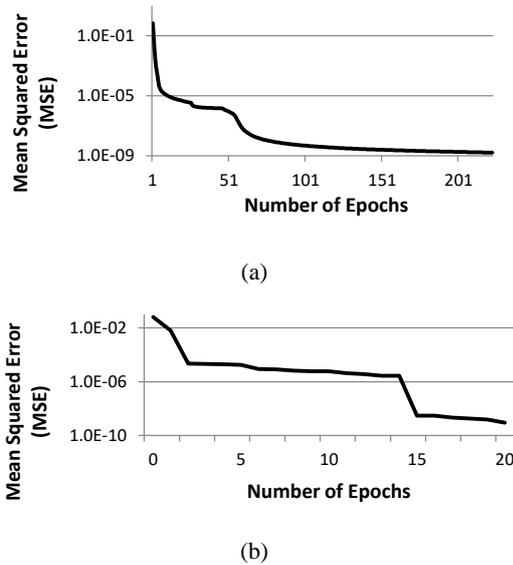

(b)

**Figure 5.** MSE of (a) MLP and (b) RBF networks based on the iterations. Both networks performed on normalized data

Performance of the network is 8.87e-10 which is an ideal value.

**4. 3. Performance on Real Data**　　By observing the satisfactory performance of both networks on normalized data, it is expected that the network can show an acceptable performance also on real data. Figure 6 represents the performance of the MLP and RBF networks for real data.

MSE of MLP network has a sharp decrease until $10^{th}$ epoch. Then, continues to decrease by a smaller slope. The best validation performance occurs at the $222^{nd}$ epoch, which is the point that the network stops training. Final MSE is 1.98e-5 which is smaller than goal error, so it can be interpreted as an ideal performance.

The performance of the RBF network decreases with an appropriate slope until $14^{th}$ epoch. At that point, the MSE value has an amount of 8.8e-5 which satisfies the goal error. In this particular case, adding the next 6 neurons does not make any change to performance and also the structure of the network.

**4. 4. Error Distribution**　　Figure 7 represents the error (Target-Output) histogram of four different simulations. Figures 7(a) and 7(b) represent the error histograms of the RBF network for normalized and real data, respectively. The distributions of the errors in both cases have an acceptable form. In better worlds, most of the errors are less than 1e-5 for normalized data and 1e-2 for real data (with a very few errors more than 1e-4 for normalized data and 1e-2 for real data).

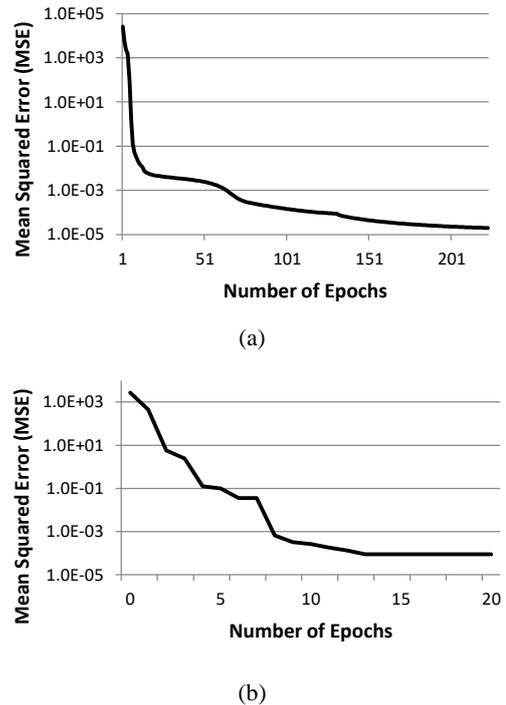

(b)

**Figure 6.** MSE of (a) MLP and (b) RBF networks based on the iterations (epochs). Both networks performed on real data

In addition, the distribution of errors is like the normal distribution, which balances out the positive and negative errors. Consequently, this distribution of errors shows that the performance of the network is acceptable.

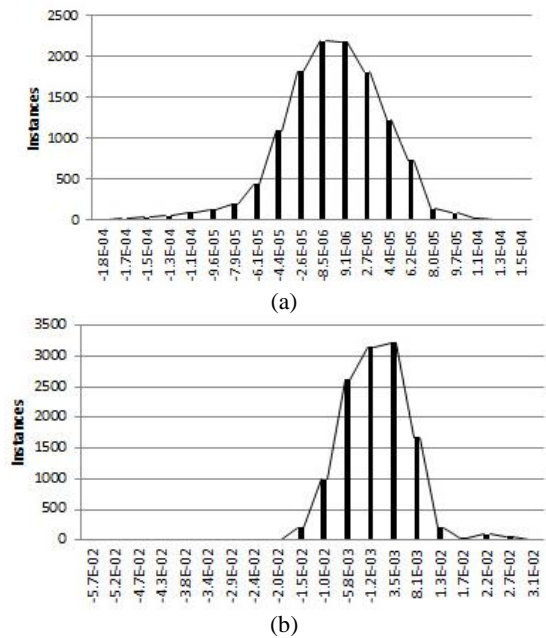

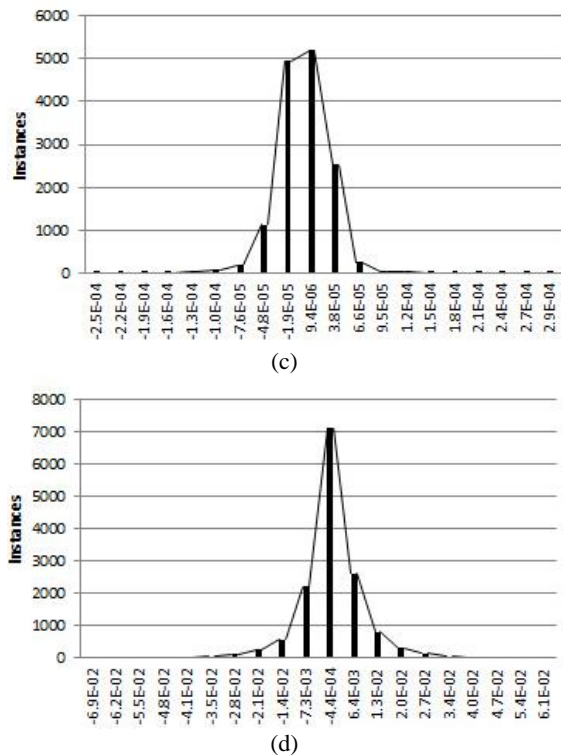

(c)

(d)

**Figure 7.** Error distribution of simulated networks

Figures 7(c) and 7(d) depict error histograms of the RBF network for normalized and real data, respectively. Similarly, distribution of errors in RBF network is also like a standard distribution. More than 90% of errors have a value less than 1e-4 for normalized data and 1e-2 for real data. This distribution shows that the network works efficiently for both normalized and real data.

## 5. CONCLUSION

Neural networks have a vast application in function approximation and optimization calculations. In this paper, two methods are introduced to simulate the inverse kinematics of a parallel robot; one of them is based on the MLP neural network and the other one is based on the RBF neural network. Both MLP and RBF approaches are successful to simulate the calculations in an acceptable time with an acceptable error rate. Implementing each of these two methods on the Tricept robot, will cause the robot work more efficiently, in respect to accuracy and reliability.

As depicted in Sections 4.2 and 4.3, proposed models can competently surplus the goal error. These sections provide two software models of the PKM Tricept, which can estimate the function of robot considerably faster than robot's normal kinematic calculation. Consequently, these types of computational intelligence models are very useful in robot controller design. In better words, the controllers need immediate feedback for appropriate functioning, but normal operation of the robot is not fast enough to satisfy these limits. The computer models like the two neural network models explained in this paper can be used to eliminate such limitations. Accordingly, analogous approaches with help of neural network methods can be useful to help many other robotic and intelligent systems to perform more effective.

# Kinematic Synthesis of Parallel Manipulator via Neural Network Approach

J. Ghasemi, R. Moradinezhad, M. A. Hosseini

*Faculty of Engineering & Technology, University of Mazandaran, Babolsar, Iran*

چکیده

در این تحقیق، شبکه های عصبی به عنوان یک ابزار قدرتمند برای حل معادلات سینماتیک معکوس یک ربات موازی به کار گرفته شده است. به این منظور معادلات سینماتیک یک مکانیزم Tricept با دو درجه آزادی چرخشی و یک درجه آزادی انتقالی توسعه داده شد. با استفاده از روش های آنالیز تحلیلی، معادلات سینماتیک معکوس برای یک فضای مشخص حل شد. پاسخ به دست آمده به عنوان ورودی شبکه عصبی به کار گرفته شده است. نتایج دو شبکه عصبی بکار گرفته شده (پرسپترون چند لایه و شبکه عصبی تابع شعاعی) با سرعت و دقت مناسبی توانست معادلات پیچیده سینماتیکی مکانیزم را مدل کرده و حل نماید.

*doi*: 10.5829/ije.2017.30.09c.04